\documentclass[paper = a4, onecloumn, 10pt]{article}

\usepackage[british]{babel}
\usepackage[utf8]{inputenc}
\usepackage{latexsym}
\usepackage[T1]{fontenc}

\usepackage{mathptmx}% http://ctan.org/pkg/mathptmx
\usepackage{amsmath}
\usepackage{mathtools}
\usepackage{titlesec}
\titleformat*{\section}{\fontsize{12pt}{0pt}\bfseries}
\titleformat*{\subsection}{\fontsize{12pt}{0pt}\bfseries}
\titleformat*{\subsubsection}{\fontsize{12pt}{0pt}\bfseries}
\titlespacing\section{0pt}{16pt plus 0pt minus 0pt}{3pt plus 0pt minus 0pt}
\titlespacing\subsection{0pt}{9.5pt plus 0pt minus 0pt}{3pt plus 0pt minus 0pt}
\titlespacing\subsubsection{0pt}{9.5pt plus 0pt minus 0pt}{3pt plus 0pt minus 0pt}
\titlespacing\paragraph{0pt}{12.1pt plus 0pt minus 0pt}{0pt plus 0pt minus 0pt}
\newcommand{\myparagraph}[1]{\paragraph{#1}\mbox{}\\}
\usepackage{geometry}
\usepackage{fancyhdr}
\usepackage{color}
\usepackage{graphicx}
\usepackage[belowskip=-5pt,aboveskip=0pt,skip=6pt,labelfont=normalfont,textfont=normalfont]{caption}
\usepackage{hyperref}
\usepackage{pbox}
\definecolor{pdflink}{rgb}{0.0, 0.0, 1.0}
\definecolor{pdfcite}{rgb}{0.0, 1.0, 0.0}
\definecolor{pdfwww}{rgb}{1.0, 0.0, 0.0}
\usepackage{cite}
\usepackage[square,numbers,sort&compress]{natbib}

\let\OLDthebibliography\thebibliography
\renewcommand\thebibliography[1]{
	\OLDthebibliography{#1}
	\setlength{\parskip}{0pt}
	\setlength{\itemsep}{0pt plus 0.3ex}
}
\usepackage{float}
\usepackage{xcolor}
\usepackage{booktabs}
\usepackage{amsmath}
\usepackage{amsfonts}
\usepackage{subcaption} % to enable subfigures
\begin{document}
	\setlength{\parindent}{0pt}
	\DeclareGraphicsExtensions{.pdf,.png,.jpg,.svg}
	
	\begin{center}  
		{\fontsize{14pt}{0pt}\textbf{Towards Reducing Data Acquisition and Labeling for Defect Detection using Simulated Data}}
        
		\vspace{19.25pt}
		$\text{Lukas Malte Kemeter}^{1}$, $\text{Rasmus Hvingelby}^{1}$,  $\text{Paulina Sierak}^{1}$,
        $\text{Tobias Schön}^{2}$, $\text{Bishwajit Gosswami}^{2}$
		\vspace{19.25pt} 
  
		$\prescript{1}{}{\text{Center for Applied Research on Supply Chain Services of the Fraunhofer Institute for Integrated Circuits IIS}}$, Nordostpark 84, 90411 Nürnberg, Germany, e-mail: malte.kemeter@iis.fraunhofer.de \vspace{19.25pt} 
  
        $\prescript{2} {}{\text{Fraunhofer Development Center X-Ray Technology EZRT, a division of the Fraunhofer-Institute for Integrated Circuits IIS}}$, Flugplatzstraße 75, 90768 Fürth, Germany, e-mail: tobias.schoen@iis.fraunhofer.de \\ 
	\end{center}

\myparagraph{Abstract} 

In many manufacturing settings, annotating data for machine learning and computer vision is costly, but synthetic data can be generated at significantly lower cost. Substituting the real-world data with synthetic data is therefore appealing for many machine learning applications that require large amounts of training data. However, relying solely on synthetic data is frequently inadequate for effectively training models that perform well on real-world data, primarily due to domain shifts between the synthetic and real-world data. We discuss approaches for dealing with such a domain shift when detecting defects in X-ray scans of aluminium wheels. Using both simulated and real-world X-ray images, we train an object detection model with different strategies to identify the training approach that generates the best detection results while minimising the demand for annotated real-world training samples. Our preliminary findings suggest that the sim-2-real domain adaptation approach is more cost-efficient than a fully supervised oracle – if the total number of available annotated samples is fixed. Given a certain number of labeled real-world samples, training on a mix of synthetic and unlabeled real-world data achieved comparable or even better detection results at significantly lower cost. We argue that future research into the cost-efficiency of different training strategies is important for a better understanding of how to allocate budget in applied machine learning projects.

\paragraph{Keywords: Domain Adaptation, Object Detection, Defect Detection, Semi-supervised Learning, Unsupervised-Learning, Cost-Efficiency}

\section{Introduction} 
\bigskip
Developing successful machine learning applications can be especially challenging when the costs for annotating data for the task of interest is high.  
The creation of a large dataset for fully supervised training is then often unreasonably expensive. In many situations, annotated data from a similar domain might be available at significantly lower costs. In our case, synthetic data is available in the form of simulated X-ray images. Following the domain adaptation literature, we refer to the available synthetic data as the \textit{source domain} and refer to the real-world data as the \textit{target domain} \cite{dshift}. \newline
When there is more source data available than target data, it is desirable to train a machine learning model only on the available source data under full supervision and apply the resulting model on the target domain as. A model trained on synthetic data alone might perform reasonably well – when evaluated on synthetic data. However, this can result in a drop in performance when the same model is evaluated on samples from the real-world domain, which it has not previously seen during training.
This drop in performance is caused by a shift in the data distribution between the source and target data and is in the following referred to as the domain gap. The domain shift problem will in many projects raise the question, whether to focus on gathering additional data annotations, synthetic data generation or more complex machine learning methods. 
This paper explores the most cost-efficient approach to mitigate this performance drop by analysing different training approaches, their demand for labeled data and their corresponding performance.  
\bigskip \newline
Section 2 presents related work, while section 3 introduces the project on which this report is based. Section 4 outlines the methodological approach for this work. Section 5 presents preliminary results, section 6 discusses future work and section 7 concludes.

\section{Related Work}

\textbf{Object detection for defect detection:} Object detection is a well studied task and has been applied to many different problems \cite{ODsurvey}. In this paper, we focus on object detection specifically for defect detection \cite{survey, surveydefects}. In defect detection, models are trained on data which usually provides bounding box information pointing to the location of a defect in an image. For making predictions, a model has to predict the location of an object (bounding box coordinates) and classify the recognised object. Established object detection architectures for defect detection are, among others \cite{ODsurvey}, the FasterRCNN \cite{FasterRCNN, FRCNNuse}, Single Shot MuliBox Detectors (SSD) \cite{SSD, SSDuse}, or YOLO \cite{YOLO, YOLOuse}. Note that we do not focus on segmentation which is also used for defect detection \cite{surveydefects}. Object detection approaches have previously shown promising results for detecting defects in X-ray projections, which is the focus of this work \cite{ren2019, FergusonAk, hu2020, mery2021, capsule}. \bigskip \newline
\textbf{Domain adaptation:} It is a well known problem that the performance of machine learning models can suffer from a domain shift problem, when models are transferred to data that differs from the original training data. Differences in the distributions of training and inference data can cause significant performance gaps \cite{dshift}. As a  solution to the domain shift problem, various domain adaptation approaches have been developed for the task of object detection \cite{survey2, DAFasterRCNN, LiuUBMT, LiCDMT}. Domain shifts can occur due to changes in the production environment \cite{tire}, cross-machine transfers \cite{crossmachine} or happen by design, when synthetic data is used to substitute expensive real-world data. Synthetic data has been used in combination with domain adaptation methods for defect detection on X-ray images \cite{tire} or other nondestructive evaluation (NDE) problems \cite{pyle2022}. The simulation to real world transfer has also been studied for other domains such as autonomous driving \cite{simtoreal} or robotics \cite{robosim2real}. However, a holistic view of the trade-off between annotation costs, improvements in synthetic data generation and domain adaptation techniques is less studied.

\section{Setup}
We consider a quality assurance system in the automotive industry, where an object detection model is developed with the goal to detect defects in X-ray scans of aluminium wheels given a limited budget. 
Acquiring and annotating real-world data for this specific use-case is expensive. Each data point corresponds to an X-ray scan of a wheel, and annotating it involves marking the defects with bounding boxes manually. For obtaining annotated data, access to defective wheels and expert knowledge is needed, which is a significant challenge. 

\subsection{Data}
\begin{figure}[h] 
		\centering
    \includegraphics[scale=0.2]{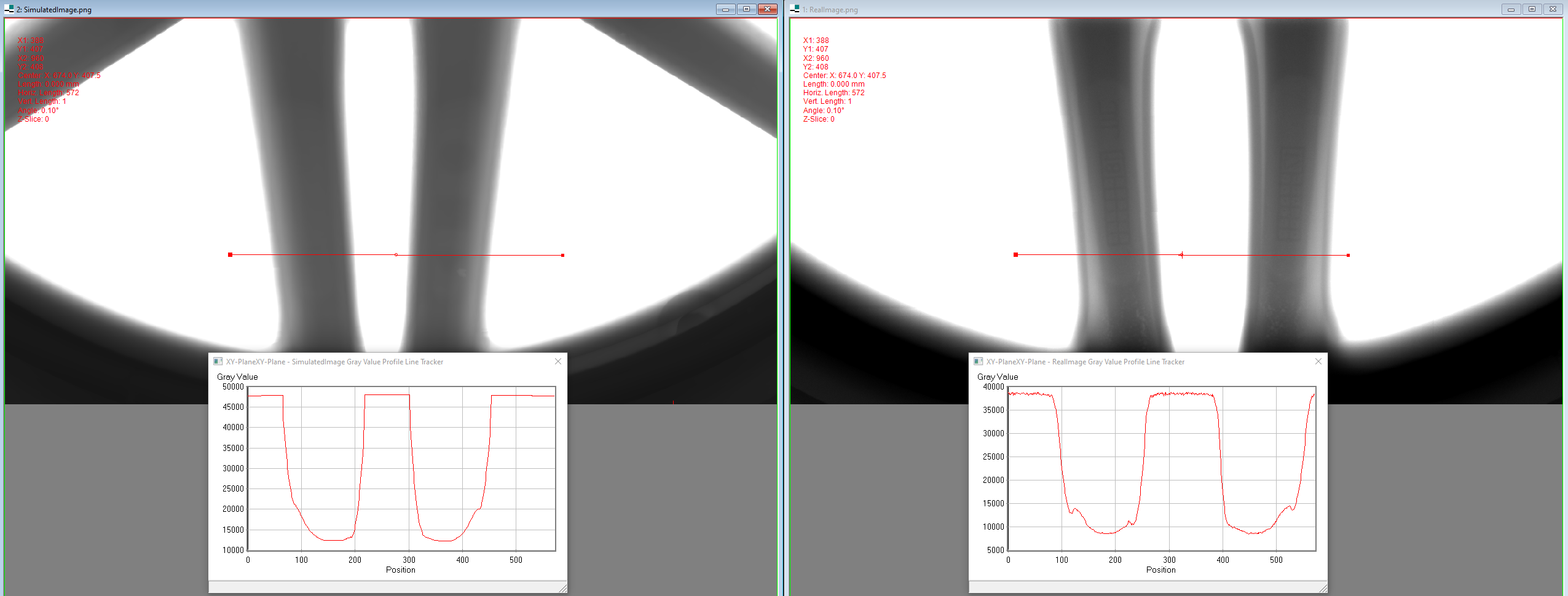}		
		\caption{A comparison between simulated (left picture) and real (right picture) X-ray projection from \protect\cite{schoen}. The shift between the source and target domain distribution is here illustrated by the grey value profile (red line) as well as by the slightly different position of the spokes on the images. The simulated images appear to have a very smooth profile, real world images appear to be more fuzzy.}
		\label{fig:appendixf1}
\end{figure}

In the first phase of the project, a simulation pipeline was developed to create synthetic X-ray projections including realistic defects and their corresponding bounding boxes. Aside from the initial development cost for the simulation tool, the marginal cost of generating an additional \textbf{annotated} image from the source domain is negligible. This synthetic data enables us to train the neural network with a substantial amount of annotated samples, which would otherwise be difficult to obtain in the real world. For a detailed description of the simulation and data collection process as well as the first preliminary empirical results we refer to Schön et al. \cite{schoen}. Figure \ref{fig:appendixf1} provides examples of simulated and real-world images. In total, \textbf{2.386} images containing more than 12.000 defects from the real world are available with annotations validated by in-house domain experts\footnote{Note that none of our images was defects-free.}. An additional set of \textbf{8.500} synthetic images containing over 35.000 defects was generated with the simulation tool. Note that multiple sets of the synthetic data were created with different levels of defect augmentation. For the results presented in this paper, only the most advanced simulation package was used. 

\subsection{Transferability from simulation to real-world}
In previous experiments we found that using only the simulated data for model training is not sufficient for detecting defects in real-world data. A model trained exclusively on annotated synthetic images did not exceed an average recall of \textbf{11\%}\footnote{Average Recall is in \cite{schoen} defined as the average across different recall scores computed with different IOU thresholds.} on the real-world test data. It is worth noting that the same model reached over \textbf{90\%} average recall when tested on simulated source data. The domain gap in our case was thus around 80 percentage points on the metric of interest. Based on these experiments two options for improving the models are present: (1) collect additional annotated real-world images or (2) explore alternative ways of leveraging the existing labels more efficiently to improve the model performance on real-world data. This poses a trade-off between maximising model performance while limiting the cost of data collection and annotation to a minimum. The objective of this paper is therefore to study the relationship between label-demand and detection performance for different training approaches. We show that the detection performance of a fully supervised oracle could have been matched by other methods with a lower demand (and thus costs) for labeled real-world data.

\section{Methodology}

\begin{figure}[H] 
		\centering
			\includegraphics[width=0.6\textwidth]{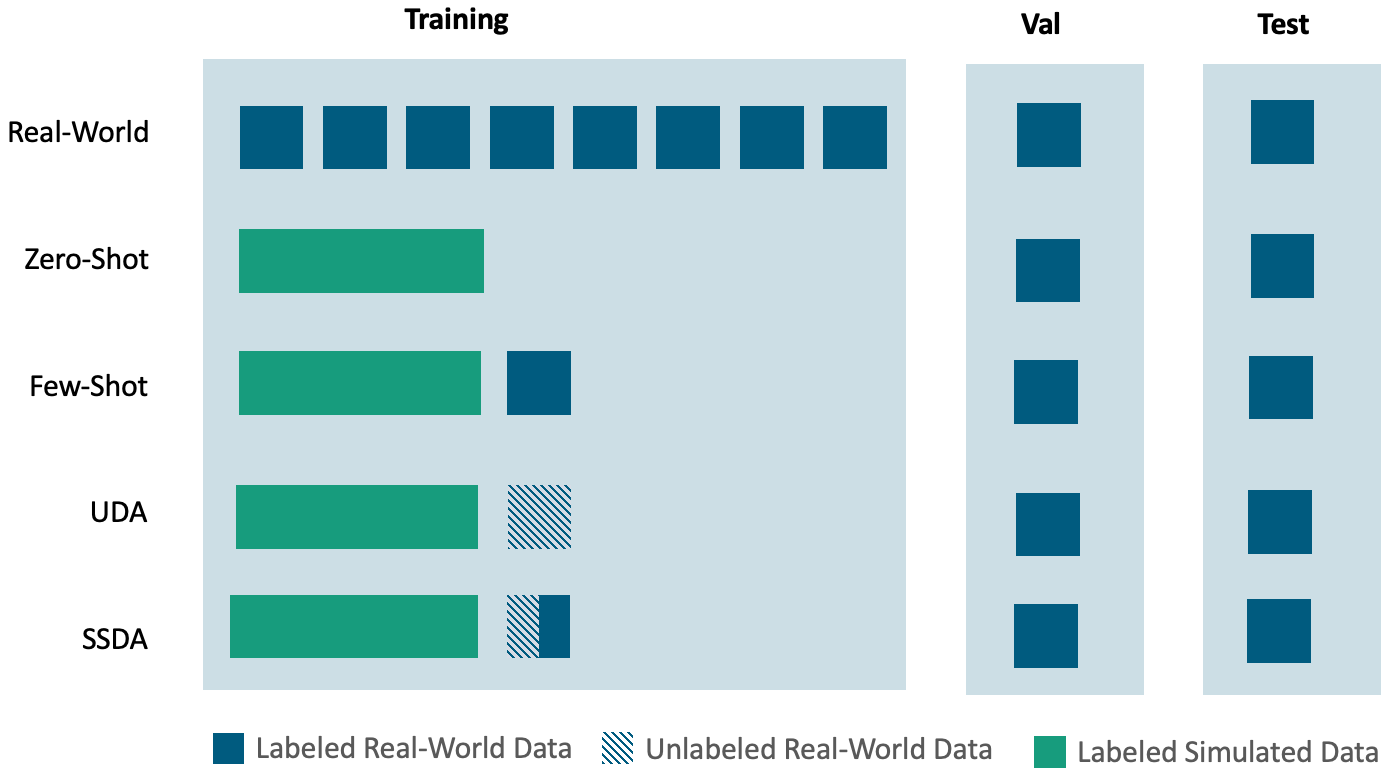}		
		\caption{Overview over experimental setup}
		\label{fig:fig1}
\end{figure}

While annotating an X-ray scan is expensive, scans without bounding boxes are relatively easy to acquire. We investigate if the demand for labeled real-world data of our models can be reduced by using both unlabeled real-world and synthetic samples in the training process. 
Figure \ref{fig:fig1} illustrates the experimental setup. As indicated by the dark blue squares, the available annotated real-world data is split into 10 folds containing 477 images per fold. Three folds accounting for 30\% of the available data will be used as a designated evaluation set. We are now interested in comparing a fully supervised model (\textit{Real-World}), a model for unsupervised domain adaptation \textit{(UDA)} and a model for semi-supervised domain adaptation (\textit{SSDA}). 

\subsection{Domain Adaptation Architecture}
In this paper, we implement a domain adaptive Faster R-CNN model \cite{DAFasterRCNN} for both the \textit{UDA} as well as the \textit{SSDA} approach. By using an adversarial training strategy based on Gradient Reversal Layers \cite{GRL}, the DA Faster R-CNN can leverage both the synthetic and unlabeled real-world data during training. When training the \textit{UDA} models, each loaded batch contains synthetic and unlabeled real-world samples.
Following Chen et al. \cite{DAFasterRCNN}, we use both the unlabeled source and target data to train a domain discriminator that learns to classify if an image is from the synthetic or real-world domain. During training, the network parameters are optimised in a way that maximises the domain classification loss, both on the image and instance level. This way, we can train a model that learns to detect defects but that at the same time is unable to differentiate between the two domains. A similar approach is taken when training the \textit{SSDA} Faster R-CNN. Additionally to the synthetic data, each batch contains both labeled and unlabeled real-world samples. The few available labeled real-world samples are used to optimise the supervised classification loss component. This is expected to improve the learning process compared to the unsupervised adaptation. For details, the reader is referred to Ren et al. \cite{FasterRCNN} and Chen et al. \cite{DAFasterRCNN}. While this technique has been shown to align the domains, it will most likely not completely close the domain gap. 

\subsection{Training Approach for Supervised Learning, UDA, and SSDA}
The major difference when training the three approaches is how the available target domain data is used. The supervised model is exclusively trained on annotated real-world samples. It represents the most cost-intensive training approach. \textit{UDA} on the other hand is trained by adding one or multiple folds of the real-world data – without labels – to the synthetic data\footnote{While annotations are available for 100\% of the real-world samples, we ignore the annotation information here to obtain unlabeled data from the target domain.}. \textit{UDA} is the approach with the lowest associated annotation costs. At last, \textit{SSDA} is trained on the synthetic data, plus a mix of labeled and unlabeled real-world data. \textit{SSDA} thus represents a compromise between full supervision at maximal costs and no supervision at minimal costs for data annotation. All three approaches are evaluated on the same splits of real-world data to ensure a fair comparison. Our strategy is now to train a model using each one of these three approaches multiple times with varying amounts of real-world data to be used in training. More specifically: As 30\% of the target data is reserved for evaluation, we can train each of the discussed approaches 7 times, using either 1 fold (10\%), 2 folds (20\%), ..., or 7 folds (70\%) of the real-world data for training. When training the \textit{SSDA} models, annotations are kept for only \textbf{1\%} of the used real-world data, the rest will be treated as unlabeled. Note that we will always use 100\% of the available labeled simulation data for training \textit{UDA} and \textit{SSDA}.

\section{Preliminary results}

\begin{figure}[h] 
		\centering
			\includegraphics[width=0.7\textwidth]{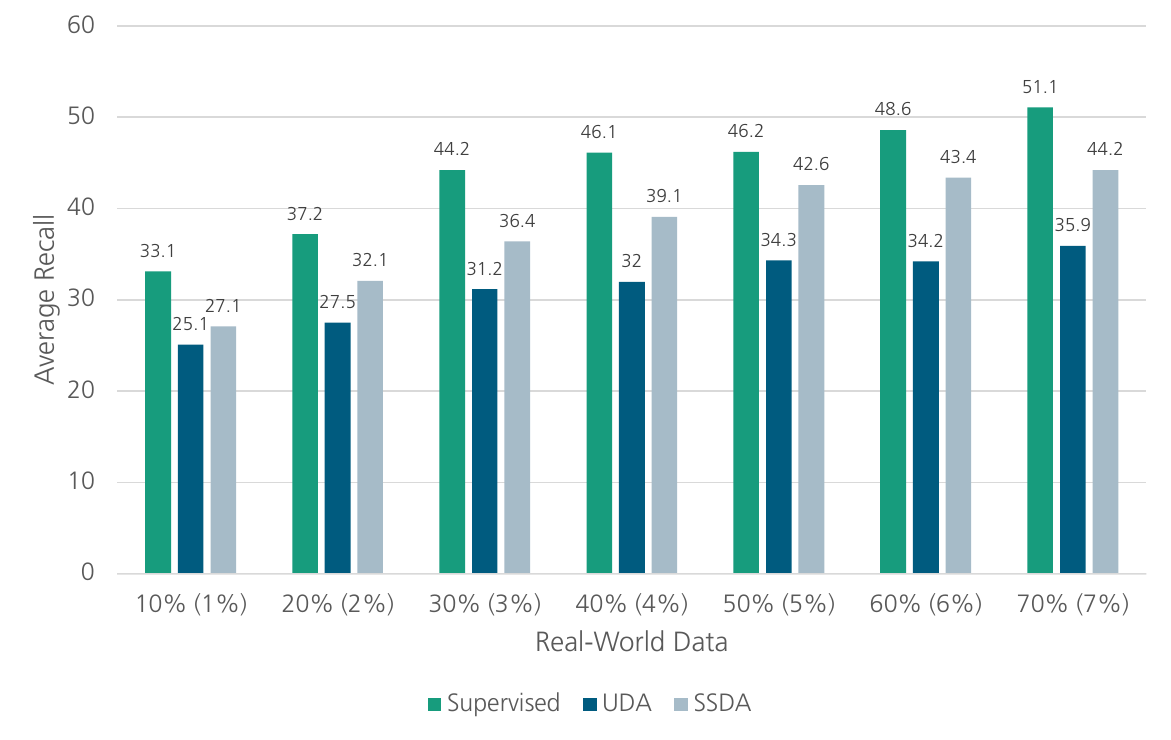}		
		\caption{Comparing supervised training with SSDA, UDA for different levels of annotated target data}
	\label{fig:compariosn_results}
\end{figure}

Figure \ref{fig:compariosn_results} compares the training results from the \textit{Supervised, UDA} and \textit{SSDA} approach. The values reported on the y-axis for each bar correspond to the average recall calculated on the evaluation folds. More specifically, it is calculated by averaging the recalls given different IOU thresholds – namely @20, @30, @40, and @50.
The x-axis tracks how much real-world data was used for training. The number in the brackets shows the share of the used real-world data that is labeled when training \textit{SSDA}. An x-axis value of \textbf{10\% (1\%)} thus implies that the green bar represents a supervised model trained with \textbf{1 fold} of the real-world data. The blue bar for \textit{UDA} results from training on synthetic data plus \textbf{1 fold} of the real-world data, without any labels. The grey bar for \textit{SSDA} results from training with synthetic data plus \textbf{1 fold} of the real-world data, of which only \textbf{1 \%} is labeled. 

Our results shows that given a fixed amount of available real-world data, full supervision consistently outperforms the two other approaches. \textit{SSDA} also consistently beats the fully unsupervised \textit{UDA} approach. Not surprisingly, a higher absolute number of labeled real-world samples in training leads to higher detection performance. But the key question here is, what is the maximum performance that can be reached given a fixed amount of annotated real-world data? 

\subsection{Improving detection performance with fewer labels}
When the number of available labeled real-world samples is fixed at a certain level, our results show that full supervision is not the most label-efficient approach. Considering the case where only 1 labeled fold is available for training and simulation of data is possible, then the green bar to the left is now the only achievable supervised model and thus our reference point. 
In this case, full supervision achieves a score of 33.1 at a fixed cost of annotating \textbf{one fold (10\%)} of real-world images (477 X-ray scans). All reported versions of \textit{UDA} and \textit{SSDA} require less annotated target data\footnote{The \textit{SSDA} model at 70\% (7\%) is the model with the highest demand for labels in our study. It requires 7 folds of real-world data but only 7\% of that is required to be labeled.} and our results show that most of them match or even outperform the supervised benchmark.

\begin{itemize}
    \item \textbf{UDA@50\%:} A similar detection performance (from 33.1 to 34.3) could have been reached without any labeled real-world data. \textbf{Cost reduction potential: }The acquisition costs for unlabeled target images increase by factor 5 while the costs for data annotation are reduced by 100\%. 
     \item \textbf{SSDA@30\%:} Performance could have been improved by roughly 10\% (from 33.1 to 36.4) by using only one third of the labeled real-world samples. 
\textbf{Cost reduction potential:} The acquisition costs for unlabeled target images increase by factor 3 while the costs for data annotation are reduced by around 66\% compared to full supervision. 
     \item \textbf{SSDA@70\%:} Following a similar line of reasoning, the detection performance could be improved by around 33\% (from 33.1 to 44.2). \textbf{Cost reduction potential:} The acquisition costs for unlabeled target images increase by factor 7 while the costs for data annotation are reduced by around 33\%. 
\end{itemize}

While domain adaptation methods can be more label- and cost-efficient than fully supervision, our results are based on the following two conditions. (1) Both \textit{UDA} and \textit{SSDA} models rely on the availability of high-quality synthetic data and (2) a high amount of available \textbf{unlabeled} real-world samples must be available. The results indicate that high quality synthetic data combined with domain adaptation methods has the potential to outperform a setup where real-world data is annotated for supervised training.

\section{Discussion}
\begin{figure}[H] 
		\centering
			\includegraphics[width=0.6\textwidth]{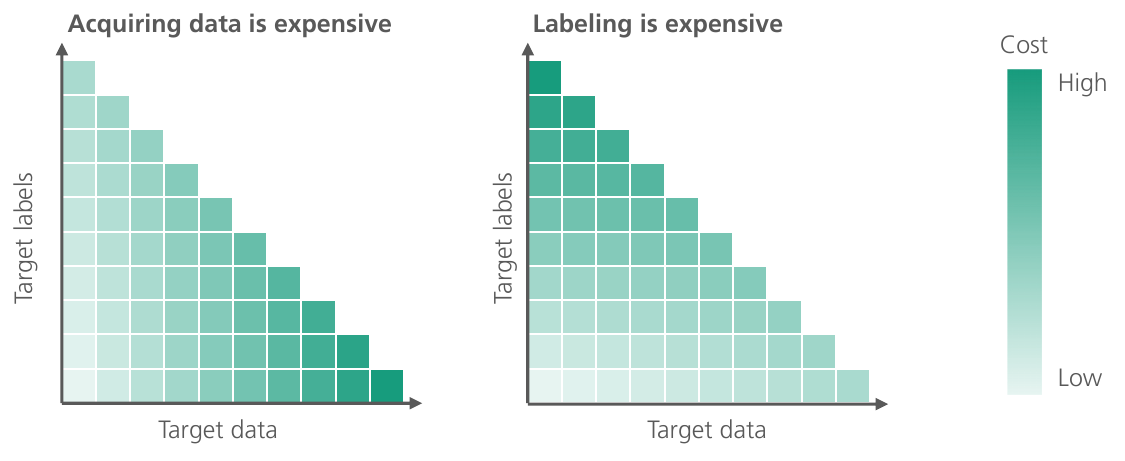}		
		\caption{Example of how the cost can vary depending on what is more expensive - acquiring data or labeling the data}
		\label{fig:cost}
\end{figure}

\begin{figure}[H] 
		\centering
			\includegraphics[width=0.6\textwidth]{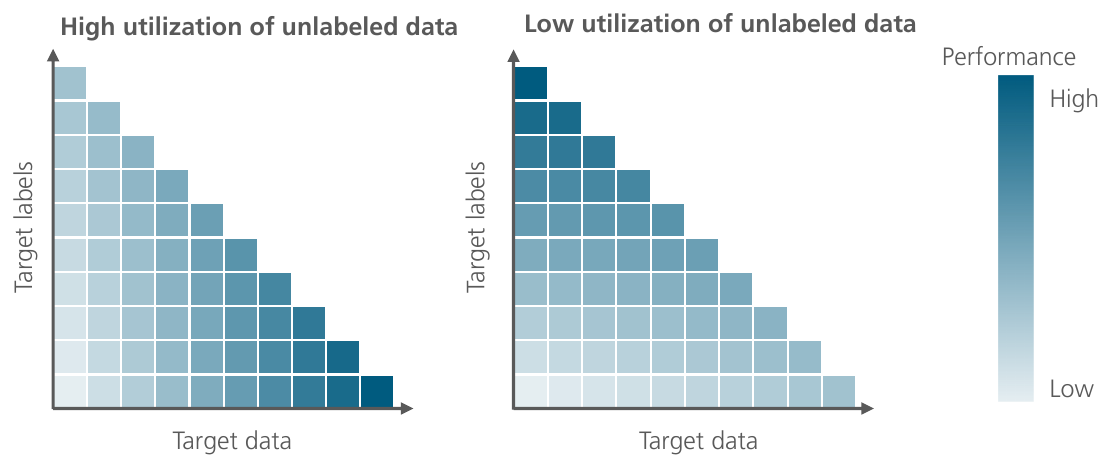}		
		\caption{Example of how performance can vary depending on the method chosen and the information available in the data}
		\label{fig:performance}
\end{figure}

We find that using simulated data in combination with domain adaptation approaches is a promising approach compared to annotating costly real-world data. In our scenario we are interested in scaling the solution to many different products where we would need to train new models per product. When we want to consider the most cost-effective solution for scaling and we can therefore think of two types of costs, namely the one-time costs for developing the system and the costs for making the solution work for new products. The latter includes the cost of compute for running the simulation or training models as well as the data collection cost and the annotation costs. Figure~\ref{fig:cost} is an example that shows how the costs can be visualised based on what the cost is for acquiring data or annotating data. 
Combining the cost information together with information on the performance of models with varying amounts of labeled and unlabeled data can provide valuable insights into how to maximize the performance while minimizing the costs. Figure~\ref{fig:performance} shows an example of how the performance of a model can vary depending on what data is available and how the data is utilized.
In future works, multiple experiments can be run for gaining insights into the most cost-efficient method in our use case. Such experiments would consist of a large number of trials with variations in the number of labeled and unlabeled data points used for training. 

As the results reported in this paper are preliminary, the underlying models can be further optimised regarding detection performance. We identify two primary approaches for enhancing performance optimisation: data-centric and model-centric improvements. 
On the model-centric approach, we are aware that while the DA Faster-RCNN can leverage the unlabeled real-world data to a certain degree, other domain adaptation methods might use the unlabeled training data in more efficient ways. Mean-teacher models have successfully been used for low-supervised domain adaptation \cite{LiuUBMT, LiCDMT} and are an interesting candidate to compare to the DA Faster R-CNN. A mean-teacher model would use a fixed teacher model to predict the unlabeled samples in training. These predicted bounding boxes and class labels can then be used to train a student model in a supervised fashion. The teacher model would be gradually updated with a moving average of the student's weights. Recently, vision transformer-based architectures have successfully been applied to object detection tasks \cite{ZhangViT, ViT, ViTObjDet} and would be another interesting architecture to investigate for our use case. Another direction that would be interesting is to explore algorithmic improvements in other fields of low-supervision such as meta-learning \cite{meta}. 

As we face a domain shift by design due to the difference between simulated and real-world images, extending the simulation pipeline is an apparent data-centric area for improvement. As simulated images and real-world images become more similar, the domain gap should decrease. We found in previous experiments that a model's performance on synthetic data improves with more sophisticated simulation approaches. By adding more variation and other augmentations such as scaling and rotations to the simulation, the resulting domain gap could be slightly reduced \cite{schoen}. Even more sophisticated adjustments to our simulation process could further mitigate this domain gap. Another area for future work could be to explore alternative ways of generating synthetic data. Generative networks have for example been used to generate synthetic X-ray images or synthetic defects for training defect detection models \cite{GANS, GANS2, GANsimdefect}.

\section{Conclusion}

We argue that a better understanding of the discussed trade-off between data annotation costs, a model's demand for annotations, and the corresponding performance is valuable – especially in situations where cheaper data from a similar domain is available. In our case, we addressed the domain shift caused by training on synthetic X-ray images by using an established domain adaptation architecture that was able to leverage information from unlabeled real-world data. As the cost of acquiring additional unlabeled X-ray samples was significantly cheaper than annotating additional real-world samples for our use case, domain adaptive training approaches proved to be more cost-efficient than a fully supervised benchmark. We show that for a given amount of available real-world X-ray scans, these methods produced better results while requiring fewer annotated real-world samples. It remains open work to quantify and generalise the monetary cost-saving potential for our and potentially other use cases in a bigger study.

\myparagraph{Acknowledgements}
The authors would like to thank the company BORBET Austria GmbH  for providing a sufficient number of wheels for scanning and tests. This work was supported by the Bavarian Ministry of Economic Affairs, Regional Development and Energy through the Center for Analytics – Data – Applications (ADA-Center) within the framework of BAYERN DIGITAL II (20-3410-2-9-8) as well as the German Federal Ministry of Education and Research (BMBF) under Grant No. 01IS18036A\\

%% Bibiliography =====================================================

\raggedright

\end{document}